%% file: ijcai24.tex
\documentclass{article}
\usepackage{ijcai24}

\usepackage{times}
\usepackage{soul}
\usepackage{url}
\usepackage[hidelinks]{hyperref}
\usepackage[utf8]{inputenc}
\usepackage[small]{caption}
\usepackage{graphicx}
\usepackage{amsmath}
\usepackage{amsthm}
\usepackage{booktabs}
\usepackage{algorithm}
\usepackage{algorithmic}
\usepackage[switch]{lineno}
\usepackage{xcolor}
\usepackage{tabularx}
\usepackage[normalem]{ulem}

\title{Faster and Lighter LLMs: A Survey on Current Challenges and Way Forward}

\author{
Arnav Chavan$^{1,2}$\and
Raghav Magazine$^{1}$ \and
Shubham Kushwaha$^{1}$ \and \\
M\'erouane Debbah$^{3}$
\And
Deepak Gupta$^2$\\
\affiliations
$^1$Nyun AI, India
$^2$Transmute AI Lab (Texmin Hub), IIT (ISM) Dhanbad, India\\
$^3$KU 6G Research Center, Khalifa University of Science and Technology, Abu Dhabi, UAE\\
\emails
arnav.chavan@nyunai.com,
guptadeepak2806@gmail.com
}

\begin{document}

\maketitle
\begin{abstract}
\vspace{-0.5em}
Despite the impressive performance of LLMs, their widespread adoption faces challenges due to substantial computational and memory requirements during inference. Recent advancements in model compression and system-level optimization methods aim to enhance LLM inference. This survey offers an overview of these methods, emphasizing recent developments. Through experiments on LLaMA(/2)-7B, we evaluate various compression techniques, providing practical insights for efficient LLM deployment in a unified setting. The empirical analysis on LLaMA(/2)-7B highlights the effectiveness of these methods. Drawing from survey insights, we identify current limitations and discuss potential future directions to improve LLM inference efficiency. We release the codebase to reproduce the results presented in this paper at \url{https://github.com/nyunAI/Faster-LLM-Survey}
\end{abstract}
\vspace{-1.5em}
\section{Introduction}
\input{sections/introduction}

\section{Model Compression: An Overview}
\input{sections/compression-overview}

\vspace{-0.5em}
\section{Experimental Analysis}

As discussed above, there exist several approaches for model compression, and there is no clear consensus on which method to use when or which method is superior over the others. Thus, we present here an experimental analysis of the different LLM compression methods and present important insights. For all the experiments, we provide practical inference metrics including model weight memory (WM), runtime memory consumption (RM), inference token rate and WikiText2 perplexity computed on a Nvidia A100 40GB GPU.

\input{pruning}
\vspace{-0.5em}
\emph{Pruning of LLaMA-7B. }In this analysis, we examine the structured pruning of the LLaMA-7B model using three recent Large Language Model (LLM) pruning methods. Table \ref{tab:pruning} showcases the performance scores for these methods at sparsity levels of 20\% and 50\%. Notably, all compression methods exhibit effective performance in terms of perplexity at lower sparsity 
 levels.Wanda-SP denotes Wanda adapted to structured pruning as reported in \cite{an2023fluctuationbased}. Noticeably, Wanda-SP and LLM-Pruner impacts the model's performance and have suboptimal results at 50\% sparsity. On the other hand, both FLaP and the fine-tuned variant of LLM-pruner perform well at this level. Comparing RM, WM, and Perplexity, these two methods demonstrate similar performance, with FLaP slightly outperforming the fine-tuning-based LLM-pruner. It is important to note that beyond superior performance, FLaP is also training-free, which makes it a preferred choice for LLM pruning.

\input{quantization}
\emph{Quantized LLaMA2-7B. }Table \ref{tab:quant} presents a comparative study demonstrating the efficacy of different quantization methods for improving LLM inference. For each quantization method choice, we default to Pytorch as the default inference engine and use propriety engines when Pytorch support is not available. As can be seen, the perplexity of all the models is mostly intact with only marginal degradation. As expected, lower precision leads to lower working and running memory consumption. Importantly, we see that at 4-bit, OmniQuant can maintain performance the most. However, GPTQ and AWQ have a wider support on different engines. Another interesting observation is that even though sub 4-bit quantizations lead to a drop in model performance, the resultant models are still better than those obtained from pruning at similar compression levels.

\input{engines}
\emph{System-level optimizations for LLaMA2-7B. }We also consider system-level optimization methods and improve LLM inference by employing various inference engines proposed in the existing literature. Related results are presented in Table \ref{tab:engines}. As can be seen, different methods have advantages across different performance metrics. TensorRT-LLM stands out with impressive performance across all metrics, particularly on NVIDIA GPUs. It provides the best token rate with GPTQ 4-bit quantization, however, efficient 4-bit support is only available for new hardware\footnote{Ampere and newer series of GPUs support 4bit runtime}. It can also be consistently seen that GPTQ is faster than AWQ at the same precision, however, the perplexity is slightly worse. Further, MLC-LLM seems to demonstrate slightly lower performance compared to TensorRT-LLM, however, its compatibility with a range of hardware positions it as a favourable choice in specific scenarios. 





\vspace{-0.5em}
\section{Challenges and Way Forward}

\emph{\textbf{Large-scale pruning/distillation is computationally intensive. }}The strategies of architecture pruning and knowledge distillation have gained widespread popularity for compressing deep learning models. However, these techniques require several fine-tuning steps, the computational demands of which can rival or even surpass the intensity of the initial training steps. In the context of LLMs, this renders them impractical, given their already substantial computational requirements. While some efforts have been made to address this challenge, they often result in significant accuracy drops even for marginal compression gains. Possible ways to circumvent the issue could include:
\begin{itemize}
    \item Revisiting the training-free pruning methods to explore their potential in the context of LLMs. For example, knowledge-preserving pruning, which focuses on reducing the unwanted knowledge context in a network rather than eliminating weights, can be improved and adapted for LLMs. Since such methods are mostly training-free, they could offer efficient LLMs at only a small additional computational budget.
    \vspace{-0.3em}
    \item Exploring layerwise pruning of LLMs. A straightforward implementation of layerwise pruning would require defining localized loss functions in terms of regression loss and compressing the sub-network while ensuring that the local output is reproduced. However, in such an approach, even small errors in the early layers could easily propagate to the later layers leading to poor performance of the compressed network.
    \vspace{-0.3em}
    \item Localized distillation of LLMs. A potential solution to overcome the issue of distillation could be to develop localized distillation methods. Instead of condensing the entire teacher LLM information into a smaller student, this approach involves learning localized parts of the teacher network in smaller-scale student sub-networks. A strategy can then be devised to combine these sub-networks into a fully compressed student LLM. This approach holds promise as a potential solution to the computational challenges associated with LLM distillation.
    \vspace{-0.3em}
    \item Growing smaller LLMs to reach the desired performance. The primary obstacle in compressing Large Language Models (LLMs) lies in the computational challenges during fine-tuning, attributed to the models' substantial size. An alternative and ambitious research direction involves growing smaller language models (SLMs) into LLMs using well-defined neural network growing strategies. This approach avoids the need to train a full-scale LLM, and the maximum computational burden is determined by the final compressed LLM obtained through the growth of the SLM.
    \vspace{-0.3em}
    \item Using PEFT methods to fine-tune when pruning efficiency. To address the challenge of full-scale fine-tuning during pruning, an alternative approach is to employ PEFT methods. Unlike traditional methods, PEFT does not require updating the model weights; only the added masks and PEFT parameters are updated \cite{zhang2023loraprune}. This significantly reduces the computational intensity of the fine-tuning process. However, PEFT methods currently face limitations in achieving large-scale compression of LLMs, indicating a need for further research to develop PEFT methods tailored specifically for compressing LLMs.
\end{itemize}

\noindent\emph{\textbf{On-the-fly Quant-Dequant makes the inference slow. }}The utilization of lower-precision floating-point formats such as FP4 poses a dual challenge regarding memory efficiency and computational speed during inference. While contemporary hardware typically supports formats like FP16 and INT8, which enable substantial memory reduction, the lower precision conversions typically needs the Quantization (Quant) and Dequantization (Dequant) operations. These operations can induce computational overhead, contributing to a slowdown in the inference process compared to using higher-precision formats like FP16. Therefore, while the adoption of lower-precision formats can offer memory efficiency gains, they adversely affect the inference speed and a right balance between the two needs to be struck.
A potential solution involves the development of streamlined Quant-Dequant operations, aiming to alleviate the observed overhead in inference speed. Another strategy is to tailor the choice of precision formats according to the specifications of the hardware in use. Concurrently, advancements on the hardware front are essential, necessitating support of lower precision formats to a broader range of popular hardwares.

\noindent\emph{\textbf{Rank selection in the layerwise low rank approximation is hard. }}
While low-rank approximation exhibits enormous potential for LLM compression, this approach is accompanied by a set of challenges, particularly in the determination of hyperparameters governing the rank reduction process. Deciding on a low-rank approximation strategy lacks a clear consensus for generalizing the method across different models. Moreover, the computational infeasibility of solving a system-level decomposition system adds a layer of complexity, making it challenging to achieve an optimal reduction in model size while preserving performance. 

It is crucial to recognize that determining the optimal rank to retain across various layers is not easily addressed through a hyperparameter search problem. Many of these approaches are computationally expensive, particularly in the context of Large Language Models (LLMs). There is a necessity to explore and develop an effective strategy for searching for the right rank when employing low-rank approximations.

\noindent\emph{\textbf{Existing evaluation metrics may not comply well. }}Compressing LLMs while preserving their ability to handle extensive contextual information is a challenge, and appropriate evaluation metrics need to be developed to tackle this issue. Another factor is the loss of fidelity. Aggressive compression may lead to a significant loss of model fidelity, impacting the language model's ability to generate accurate and contextually relevant outputs. Several such characteristics of LLMs need to be captured in their compressed variants, and this can only be identified by the right choice of metrics.

\noindent\emph{\textbf{Python - an interpreted language leads to slower execution times. }}While Python is a versatile and widely used programming language, it is inherently interpreted, which can lead to performance bottlenecks, especially in computationally intensive tasks such as deep learning. The Global Interpreter Lock (GIL) in CPython, the default Python interpreter, further restricts the concurrent execution of multiple threads, limiting the language's ability to fully exploit the potential of multi-core processors. This highlights the need to seek alternative solutions to improve the speed of deep learning workflows.

Some of these issues have been overcome with the development of optimized libraries and frameworks, such as TensorFlow and PyTorch, which incorporate high-performance kernels implemented in lower-level languages like C++ or CUDA. However, there are several other ends where Python restricts the model performance. An illustrative example is LLama.cpp, where the transition to a C++ implementation, LLaMA-7B, resulted in significantly improved speed. This shift exemplifies the impact of choosing a language optimized for performance in the context of deep learning models. Moreover, the emergence of Rust-based models has attracted attention for their superior speed. Rust, with its emphasis on both memory safety and performance, has demonstrated effectiveness in accelerating computations, particularly in scenarios where speed is of paramount importance. Thus, for optimizing the inference speed, moving away from Python to C++, Rust or other similar languages might be a future direction to pursue.

\noindent\emph{\textbf{Ethical and bias considerations not necessarily maintained. }}LLMs are initially trained on extensive datasets, ensuring that the model remains statistically unbiased towards any specific case. However, during the process of model compression, a specific dataset is typically employed. As the LLM may lose some generic characteristics irrelevant to the target dataset, there is a potential to introduce unnoticed bias through standard evaluation practices. Consequently, there is a need to develop innovative evaluation strategies to guarantee that ethical concerns and biases are minimized in the compressed LLMs produced.

\vspace{-0.5em}
\section{Conclusions}
In conclusion, our survey extensively explores LLM compression, covering both model-level and system-level efficiency enhancements. We discuss various compression methodologies and provide practical insights from experiments conducted on LLaMA(/2)-7B, offering valuable information for optimizing LLMs. Analysis of the survey and experimental results highlights the existing bottlenecks in enhancing LLM inference, indicating the necessity for further developments to achieve efficiency. We envision this survey as a stepping stone towards advancing the field and achieving the goal of efficient LLM inference.

\pagebreak
\pagebreak
\appendix

\bibliographystyle{named}
\bibliography{ijcai24}

\end{document}

%% file: sections/introduction.tex
The advent of LLMs, marked prominently by models such as GPT \cite{brown2020language} and LLaMa \cite{touvron2023llama,touvron2023llama2} series, has paved a new revolution in language-related tasks, ranging from text comprehension and summarization to language translation and generation. These models, often consisting of billions of parameters, have shown remarkable performance in capturing intricate patterns, fine-detailed contexts, and semantic representations in natural language. As a consequence, they have become indispensable tools in various applications, leading to advancements in various domains, including artificial intelligence, information retrieval, and human-computer interaction.

Despite their unparalleled performance, widespread adoption of LLMs is hindered by their substantial computational and memory requirements, which pose challenges for deployment in resource-constrained environments. For example, loading a LLaMa-70B model requires 140GB of VRAM excluding the memory required for model inferencing. The need for efficient deployment has led to recent research into model compression as well as system-level modification techniques tailored specifically for LLMs. These early works have identified potential ways to improve the inference efficiency of LLMs. However, the current improvements are often accompanied by significant drops in the performance of the model, and novel research directions need to be identified to find the desired solutions to this problem.

A recent survey study has provided a concise overview of the recently proposed LLM compression methods, as well as the evaluation metrics and the data used to benchmark them \cite{zhu2023survey}. However, to further push the frontiers of research towards practical inference improvement for LLMs, a comprehensive study is still missing. In this survey paper, we explore existing methods that aim at making LLMs efficient through model compression as well as through system-level optimizations. To fairly compare various methods, we provide empirical observations using different compression techniques applied to LLaMa(/2)-7B. Our evaluation includes methods that provide a practical advantage and include structured pruning, quantization, and system-level optimizations provided by different inference engines from the existing literature. We share valuable insights drawn from these experiments to present a useful and practical understanding of efficient LLMs. Additionally, we make the code and benchmarks associated with the experiments publicly available. We also examine the difficulties linked to current compression methods in both general deep learning and those specifically suggested for LLMs, and we discuss potential directions of research to overcome these problems.

Overall, the contributions of this paper are as follows.
\begin{itemize}
\item We offer a brief overview of the model compression domain, emphasizing essential methodologies that have made notable contributions to the field of lighter and faster LLMs.
\item Complementary to model compression, system-level modifications have played an important role in speeding up the LLM inference, and we discuss these approaches as well. 
\item To provide a practical perspective, we present an empirical analysis of well-known compression methods for LLMs under a standardized setup. The insights derived can help make informed decisions about the selection of LLM compression methods based on the deployment environment.
\item Drawing upon insights derived from our survey and empirical analysis, we systematically pinpoint existing limitations and propose viable pathways forward for achieving optimal efficiency in LLM inference.
\end{itemize} 

%% file: sections/compression-overview.tex
Model compression techniques have emerged as a crucial area of research, offering promising solutions to enhance the efficiency of resource-intensive deep learning models. The domain of developing efficient Large Language Models (LLMs) can significantly benefit from insights and methodologies used in this field. Before diving into the topic of building efficient LLMs and the existing works around it, we provide an overview of some of the popular approaches employed in deep learning model compression. Below, we first introduce the traditional approaches of model compression and briefly discuss the development related to the traditional deep learning models. Following this, we provide an overview of the works related to compression of LLMs in the existing literature. 

\subsection{Compression of Deep Models}
\emph{\textbf{Architecture pruning}} refers to the process of systematically reducing the complexity of a neural network by eliminating redundant or less impactful connections, neurons, or entire layers \cite{PhysRevA.39.6600}. This technique aims to enhance model efficiency, reduce computational costs, and mitigate overfitting without significantly compromising performance. Pruning involves identifying and removing connections or units based on various criteria, such as weight magnitudes \cite{li2016pruning}, activation patterns \cite{molchanov2016pruning}, or sensitivity analysis \cite{sanh2020movement}. The pruned model retains its critical features while achieving a more compact representation, which is particularly valuable in scenarios with limited computational resources, such as edge devices or mobile applications. 

Among the widely studied pruning methodologies, the lottery ticket hypothesis \cite{frankle2019lottery} provided fundamental insights into the impact of weight initialization and pruned network structure on neural network pruning. Network Slimming \cite{liu2017learning,chavan2022vision} introduced a method to prune channels in CNNs and reduce the size of weight dimensions in Transformers by imposing sparsity regularization on the channel scaling factor. Movement pruning demonstrated large-scale pruning of BERT \cite{kenton2019bert} models by leveraging the first-order information i.e. retain weights moving away from zero, as compared to zero-order methods which retain weights with larger magnitudes. \cite{lagunas2021block} introduced block structures in weight matrices of transformer layers and employed movement pruning on them for practical speedups. More recently, \cite{jiang2023pruning} argued that fine-tuning is redundant for first-order pruning and proposed Static Model Pruning (SMP), a fine-tuning free pruning method for language models.
 
\emph{\textbf{Quantization}} reduces the precision of numerical values in a neural network, typically from 32-bit floating-point numbers to lower bit-width representations, such as 8-bit integers thus shrinking the memory footprint of the model, accelerating inference speed, and enabling more efficient deployment on hardware with limited computational resources. During quantization, weights and/or activations are rounded off to a discrete set of values, introducing a trade-off between computational efficiency and model accuracy. Even with this reduction in precision, state-of-the-art quantization methods are capable of minimizing the impact on performance.

Quantization-Aware Training (QAT) \cite{ni2020wrapnet} involves the quantization of model parameters throughout the training process, encompassing both the forward pass and backward propagation. LSQ \cite{esser2019learned} proposed a learnable step size for each weight in conjunction with other network parameters. \cite{tailor2021degreequant} introduced an architecture agnostic method for pruning graph neural networks. On the other hand, Post Training Quantization (PTQ) \cite{banner2019posttraining} finds out the optimal clipping range and channel-bit-width settings for weights and activations. OSME \cite{choukroun2019lowbit} proposed a PTQ method in which \textit{l2}-distance between the quantized tensor and the corresponding floating-point tensor is minimized.

\emph{\textbf{Knowledge distillation}} aims at training a computationally efficient model, often referred to as the student model, to mimic the predictions of a larger and more complex model known as the teacher model. This process involves transferring the knowledge embedded in the teacher model, typically characterized by its soft probabilities or intermediate representations, to the student model. Distillation is particularly useful when deploying models in scenarios with limited computational resources, as it enables the creation of smaller models that retain the performance of their larger counterparts. Additionally, distillation helps combat issues such as over-fitting, improves generalization, and facilitates the transfer of knowledge learned by deep and complex models to simpler architectures.

Knowledge distillation techniques can be divided into three classes i.e. response-based, feature-based and instance-relation based. Response-based distillation \cite{hinton2015distilling} trains the student model to mimic the final outputs of the teacher, while feature-based distillation \cite{tian2022contrastive} trains the student to mimic intermediate feature maps of the teacher. Relation-based distillation takes one step further by using an objective which models the co-relation of the similarity between various feature maps of the student and teacher network. More recently, \cite{chen2023disco} used knowledge distillation during the pre-training stage and reduced the size of BERT by 40\%, making it 60\% faster while retaining 97\% of its language understanding abilities.

\emph{\textbf{Low-rank decomposition}} reduces the computational complexity of models by decomposing weight matrices into smaller ones with fewer dimensions which in turn approximate the initial full-rank matrix. This also reduces the number of parameters to be stored in the model and speeds up matrix multiplications hence reducing memory and latency requirements. 

\cite{jaderberg2014speeding} proposed an architecture agnostic method of accelerating convolutional layers using tensor decomposition and discriminative fine-tuning; whereas \cite{denton2014exploiting} proposed clustering methods with low-rank factorization for faster CNNs. \cite{6638949} examined low-rank matrix factorization in acoustic models, where the factorization was applied on the final layer of the network.  \cite{lebedev2015speedingup} introduced canonical polyadic decomposition which is calculated using non-linear least squares for speeding up CNNs. \cite{tai2016convolutional} proposed a global decomposition optimization algorithm and thus performed better than iterative methods. 
\subsection{Compression of LLMs}

The compression of LLMs represents a distinctive challenge compared to traditional deep learning models, primarily due to the substantial scale of the former. Many established compression methods rely on the paradigm of executing fine-tuning steps to regain lost performance during the compression stage. However, this approach encounters significant limitations when applied to LLMs owing to their considerable size, necessitating a paradigm shift in the treatment of LLM compression as an independent and new research domain.

\emph{\textbf{Architecture pruning. }}LLM-Pruner \cite{ma2023llmpruner} used Taylor series expansion by leveraging a single gradient step to estimate important parts of a pre-trained LLM. LoRAPrune \cite{zhang2023loraprune} outperformed LLM-Pruner by using gradients of LoRA \cite{hu2021lora} weights, offering computational efficiency. LoRAShear \cite{chen2023lorashear} identified dependencies in LLMs, separated trainable variables into groups, and achieved compression through pruning and fine-tuning. Sheared LLaMA \cite{xia2023sheared} introduced targeted structured pruning and dynamic batch loading for end-to-end component removal. FLaP \cite{an2023fluctuationbased} a fine-tuning free structured pruning method which used a fluctuation based metric to determine the importance score of various weight columns.

Unstructured pruning methods, such as SparseGPT \cite{frantar2023sparsegpt}, adopted a one-shot technique without the need for fine-tuning. WANDA \cite{sun2023simple} pruned weights based on the product of weight values and activation inputs, eliminating the need for fine-tuning. Another recent work suggested fusing of OBS \cite{hassibi1993optimal} and OBD \cite{lecun1989optimal} criteria for weight selection and determining layer sparsity based on sensitivities derived from Hessian matrices \cite{2023arXiv231009499S}. While the structured and unstructured methods mentioned above show promise, the observed performance drop for the achieved compression level remains relatively high. Further efforts are required in developing pruning methods that can lead to efficient LLMs.

\emph{\textbf{Quantization. }}This class of methods has been relatively more successful in the compression of LLMs. LLM.\texttt{int8()} \cite{dettmers2022llmint8} made it possible to convert the higher bit LLM weights into 8-bit without deterioration in performance post-training. They proposed a two-stage quantization scheme with vector-wise quantization and mixed-precision decomposition for outliers. SmoothQuant \cite{xiao2023smoothquant}, a training-free PTQ method, reduced both weights and activations of LLMs to 8 bits. QLoRA \cite{dettmers2023qlora} introduced 4-bit NormalFloat (NF4) and double quantization to save memory without losing out on the performance of models. OmniQuant \cite{shao2023omniquant} introduced Learnable Weight Clipping (LWC) and Learnable Equivalent Transformation(LET). LWC prevents weights from attaining extreme values by optimizing the clipping threshold, while LET deals with activation outliers by quantizing weights instead of activations through a LET. SqueezeLLM \cite{kim2023squeezellm} enabled compression up to 3-bit by using a sensitivity-based non-uniform quantization scheme, where second-order information is used to find the optimal bit precision. GPTQ \cite{frantar2023gptq} used second-order information to compress models with up to 175 billion parameters to as low as 3 bits per weight with minimal loss in accuracy, pushing the previously proposed 8-bit methods to a smaller size. \cite{lin2023awq} observed that retaining 1\% crucial weights can help reduce the degradation in quantization performance. They proposed Activation-aware Weight Quantization (AWQ) which finds the best channel-wise scaling, outperforming existing techniques in general language modeling and domain-specific tasks. ZeroQuant-FP \cite{wu2023zeroquantfp} focused on floating point quantization and found that FP8 outperforms INT8 for activations and FP4 is comparable to INT4 for weights. They also incorporated low-rank compensation into their approach for enhancement. EXL2\footnote{https://github.com/turboderp/exllamav2} proposed a mixed-precision quantization algorithm, where different precision types for each layer are computed while measuring quantization errors. Their algorithm saves all the tries and associated error rates in the measuring pass and given a target precision, the algorithm quantizes the model by choosing, for each layer's module, a target precision with the lowest error rate. GGUF/GGML\footnote{https://github.com/ggerganov/ggml} proposed a mixed set of quantizations to achieve K-Quants, a \textit{mostly K} quantization output. For example, 4bit K-Quant uses 6bit for a few of the Attention and MLP layers and the usual 4bit for others.

LLM-QAT \cite{liu2023llmqat} proposed a data-free distillation method where they queried a pre-trained model to generate data which was used to train a quantized student model using a distillation loss. With the quantization of the KV-cache as well, apart from weights and activations, they can quantize 7B, 13B, and 30M LLaMA down to 4 bits. BitNet \cite{wang2023bitnet} introduced a 1-bit LLM transformer architecture. It mainly replaces the standard \texttt{nn.Linear} in PyTorch with \texttt{BitLinear} to train 1-bit weights. As the size of the models increases, it comprehensively outperforms counterparts trained on FP16. \cite{tao2022compression} proposed token-level contrastive distillation and used dynamic scaling to make quantizers adaptive to different modules.

\emph{\textbf{Knowledge distillation. }}Among the knowledge distillation methods, both white-box as well as black-box methods have been used to compress large open-source language models. Instead of solely relying on a fixed set of output sequences, Generalized KD \cite{agarwal2023generalized} trains the student on its self-generated output sequences by leveraging feedback from the teacher on such sequences. TED \cite{liang2023more} employs a dual-stage training process. In the first stage, task-specific loss trains filters in both student and teacher models. In the second stage, the student and its filters undergo training with a task-aware layer-wise distillation loss, alongside student-teacher and task-specific losses. In another work \cite{jha2023train}, the student model is initialized with a subset of layers of the teacher and trained on the same corpus and objective as the teacher. This helps to achieve task-agnostic compression without using any distillation loss.

Other distillation methods include black-box techniques such as Lion \cite{jiang2023lion}, where the student network is trained using a three-stage adversarial loop consisting of an imitation, discrimination, and generation stage. In its discrimination stage, a propriety LLM is used to find hard instructions, i.e. instructions for which the student's outputs significantly differ from the teacher's outputs. As a final step, the propriety LLM generates more samples similar to the hard instructions on which the student is trained to complete the loop. DISCO \cite{chen2023disco} is a counterfactual knowledge approach in which a propriety LLM is given a prompt and is made to generate counterfactual augmentations in it. Then a task-specific teacher model filters out these augmentations, and the student model is trained on them. SCOTT \cite{wang2023scott} used contrastive decoding to generate rationale from the teacher along with the usual question-answer pair to train the student model.

\emph{\textbf{Low rank approximations. }}TensorGPT \cite{xu2023tensorgpt} compressed the embedding layer of LLMs through Tensor-Train Decomposition and stored it in a reduced Matrix Product State, which can be computed in a distributed fashion. LoSparse \cite{li2023losparse} approximated weight matrix in LLMs as the sum of a sparse matrix and another low-rank approximation matrix. The low-rank matrices capture the expressive features among neurons as they involve doing Singular Value Decomposition and the remaining features are captured by the sparse matrix. \cite{kaushal2023lord} show that a simple decomposition of the matrices in LLMs as a product of two sparse low-rank matrices can offer noticeable compression and speedup at a small compromise of perplexity.

Overall, the research direction of using low-rank approximations to compress LLMs is new but exhibits the potential to improve inference efficiency. Two recent works have shown that low-rank approximations can often improve reasoning abilities and undergo compression through layerwise rank reduction in the weight space \cite{sharma2023truth} and/or in the latent feature space \cite{chavan2023rethinking}. These methods offer the advantage of requiring minimal computational resources for the compression process due to their layerwise approach to matrices involved. However, it should be noted that the level of lossless compression achieved using these techniques remains modest, and further improvements are needed from a practical point of view.

\emph{\textbf{System level approaches. }}
Here we highlight those methods which improve the complementary infrastructure and runtime architecture of LLMs.

\emph{Paged Attention \cite{kwon2023efficient} -} inspired by the classical virtual memory and paging techniques in operating systems, it allows storage of continuous keys and values cached in non-contiguous memory.

\emph{Tensor Parallelism -} entails dividing a tensor into shards distributed across various GPUs, processing each shard independently and in parallel, and subsequently synchronizing the results at the end of the step.

\emph{Pipeline Parallelism -} allows a model to be vertically split across multiple GPUs at the layer level, where each GPU handles one or several layers, enabling parallel processing of distinct stages in the pipeline.

\emph{CPU/GPU Offloading \cite{song2023powerinfer}-} involves transferring specific weight layers to GPU devices for matrix multiplication, subsequently transmitting the computed results back to the secondary device (RAM), thus optimizing parallel processing capabilities while allowing the secondary device to handle the remaining memory intensive computations.

\emph{Flash Attention(/v2) \cite{dao2022flashattention,dao2023flashattention} -} optimizes attention computation by employing incremental softmax reduction through input block tiling, avoiding the need for whole-input access, and expedites the backward pass by storing the softmax normalization factor from the forward pass, eliminating the requirement to read the large attention matrix from high bandwidth memory (HBM). Building upon the advancements of FlashAttention, FlashAttention-2 minimizes non-matrix multiplication FLOPs, optimizing the online softmax technique, introducing parallelism over sequence length, and refining workload partitioning among warps within each thread block to reduce synchronization, thereby achieving optimized performance on modern GPUs.

\emph{Fused Operations -} involves consolidating multiple computational tasks, such as combining existing kernels or creating new ones, to minimize the overhead associated with multiple kernel API invocations.

\emph{Speculative Decoding \cite{leviathan2023fast}-} efficiently generates multiple future tokens from a chosen smaller model and verifies them in parallel using the larger model, enabling the simultaneous decoding of multiple tokens per step.

Notable implementations in this category include vLLM\footnote{https://github.com/vllm-project/vllm}\cite{kwon2023efficient}, Llama.cpp\footnote{https://github.com/ggerganov/llama.cpp}, ExLlama(/v2), TensorRT-LLM\footnote{https://github.com/NVIDIA/TensorRT-LLM}, MLC-LLM\footnote{https://github.com/mlc-ai/mlc-llm}, PowerInfer\footnote{https://github.com/SJTU-IPADS/PowerInfer} \cite{song2023powerinfer}, among others. vLLM employs paged attention through a KV-Cache manager that separates logical and physical KV blocks, enabling dynamic growth of the KV cache. ExLlama(/v2) implements fused kernels to minimize launch overheads and API invocation overheads when operating on discontinuous blocks. Llama.cpp is a low-level C/C++ implementation of the LLaMA architecture with support for multiple BLAS backends for fast processing. It operates on the GGUF quantization scheme with CPU and GPU offloading. MLC-LLM focuses on compiler accelerations and runtime optimizations for native deployment across platforms. It encapsulates model execution logic in a container - Intermediate Representation Module (IRModule) which captures the hierarchical structure of computations for optimization and code generation. It employs Paged Attention, Fused Operators, and automatic generation of optimized kernel code for multiple hardware platforms. TensorRT-LLM implements masked multi-head attention with on-the-fly pre-processing of QKV elements. It supports Paged Attention, INT8/FP8 caches, in-flight batching, and tensor/pipeline parallelism for speedups. An additional improvement is attained due to fused in-flight batching with operation fusion. PowerInfer adopts a GPU-CPU hybrid approach, by pre-loading consistently activated hot neurons onto the GPU for fast access, computing variable cold neurons on the CPU, and integrating adaptive predictors and neuron-aware sparse operators to optimize efficiency.

Overall, these methods work complementary to model compression methods and improve the runtime efficiency of large language models. These engines demonstrate the feasibility and benefits of optimizing the software architecture and infrastructure complementary to model compression.

%% file: pruning.tex
\begin{table}[htb!]
\centering
\vspace{-0.5em}
\caption{Performance measures for various compressed variants of LLaMA-7B model obtained using the following structured pruning methods: Wanda-SP, LLM-pruner and FLaP. Here, $*$ refers to a fine-tuned variant of LLM-pruner.}
\vspace{-0.5em}
\resizebox{0.48\textwidth}{!}{
\begin{tabular}{@{}lrrrrr@{}}
\toprule
\textbf{Method} & \textbf{Sparsity} & \textbf{RM (GB)} & \textbf{WM (GB)} & \textbf{Tokens/s} & \textbf{Perplexity} \\ \midrule
Baseline & - & 26.16 & 12.55 & 30.90 & 12.62 \\ \hline
Wanda-SP & 20\% & - & - & - & 22.12 \\
& 50\% & - & - & - &  366.43\\
LLM-Pruner & 20\% & 10.38 & 10.09 & 32.57 & 19.77 \\
& 50\% & 6.54 & 6.23 & 40.95 & 112.44 \\
LLM-Pruner* & 20\% & 10.38 & 10.09 & 32.57 & 17.37 \\
& 50\% & 6.54 & 6.23 & 40.95 & 38.12 \\
FLaP & 20\% & 9.72 & 9.44 & 33.90 & 14.62 \\
& 50\% & 6.26 & 6.07 & 42.88 & 31.80 \\ \bottomrule
\end{tabular}
}
\label{tab:pruning}
\vspace{-0.5em}
\end{table}

%% file: quantization.tex
\begin{table*}[ht]
\centering
\caption{Performance comparison of different quantization methods for the compression of LLaMA2-7B. Here, WM and RM refer to weight memory and running memory consumption, respectively.}
\vspace{-0.5em}
\resizebox{0.7\textwidth}{!}{
\begin{tabular}{@{}llrrrr@{}}
\toprule
\textbf{Method} & \textbf{Inference Engine} & \textbf{WM (GB)} & \textbf{RM (GB)} & \textbf{Tokens/s} & \textbf{Perplexity} \\
\midrule
Baseline FP16 & PyTorch & 12.55 & 26.16 & 30.90 & 5.85 \\
GPTQ 2bit & PyTorch & 2.11 & 2.98 & 20.91 & NaN \\
GPTQ 3bit & PyTorch & 2.87 & 3.86 & 21.24 & 7.36 \\
GPTQ 4bit & PyTorch & 3.63 & 4.65 & 21.63 & 6.08 \\
GPTQ 8bit & PyTorch & 6.67 & 7.62 & 21.36 & 5.86 \\
AWQ 4bit GEMM & PyTorch & 3.68 & 4.64 & 28.51 & 6.02 \\
AWQ 4bit GEMV & PyTorch & 3.68 & 4.64 & 31.81 & 6.02 \\
QLoRA (NF4) & PyTorch & 3.56 & 4.84 & 19.70 & 6.02 \\
LLM.int8() & PyTorch & 6.58 & 7.71 & 5.24 & 5.89 \\
K-Quants 4bit & Llama.cpp & 3.80 & 7.38 & 104.45 & 5.96 \\
OmniQuant 3bit & MLC-LLM & 3.20 & 5.10 & 83.4 & 6.65 \\
OmniQuant 4bit & MLC-LLM & 3.80 & 5.70 & 134.2 & 5.97 \\
\bottomrule
\end{tabular}}
\label{tab:quant}
\vspace{-1.5em}
\end{table*}

%% file: engines.tex
\begin{table*}[ht]
\centering
\caption{Performance comparison of compressed variants of LLaMA2-7B using various inference engines, quantized for different predictions and across different hardwares. Here, WM and RM denote weight memory and running memory consumption respectively.}
\vspace{-0.5em}
\resizebox{0.9\textwidth}{!}{\begin{tabular}{@{}lllrrrr@{}}
\toprule
\textbf{Method} &\textbf{Hardware Support} & \textbf{Quantization Type} & \textbf{WM (GB)} & \textbf{RM (GB)} & \textbf{Tokens/sec} & \textbf{Perplexity} \\ \midrule
Llama.cpp & NVIDIA GPU & GGUF K-Quant 2bit & 2.36 & 3.69 & 102.15 & 6.96 \\
& AMD GPU & GGUF 4bit & 3.56 & 4.88 & 128.97 & 5.96 \\
& Apple Silicon & GGUF AWQ 4bit & 3.56 & 4.88 & 129.25 & 5.91 \\
& CPU & GGUF K-Quant 4bit & 3.59 & 4.90 & 109.72 & 5.87 \\
& & GGUF 8bit & 6.67 & 7.78 & 93.39 & 5.79 \\
& & GGUF FP16 & 12.55 & 13.22 & 66.81 & 5.79 \\
\hline
ExLlama & NVIDIA GPU & GPTQ 4bit & 3.63 & 5.35 & 77.10 & 6.08\\ 
& AMD GPU & & & & & \\ \hline
ExLlamav2 & NVIDIA GPU & EXL2 2bit & 2.01 & 5.21 & 153.75 & 20.21 \\
& AMD GPU & EXL2 4bit & 3.36 & 6.61 & 131.68 & 6.12 \\
& & GPTQ 4bit & 3.63 & 6.93 & 151.30 & 6.03 \\
& & EXL2 8bit & 6.37 & 9.47 & 115.81 & 5.76 \\
& & FP16 & 12.55 & 15.09 & 67.70 & 5.73 \\ \hline
vLLM & NVIDIA GPU & AWQ GEMM 4bit & 3.62 & 34.55 & 114.43 & 6.02 \\
& AMD GPU & GPTQ 4bit & 3.63 & 36.51 & 172.88 & 6.08 \\
& & FP16 & 12.55 & 35.92 & 79.74 & 5.85 \\ \hline
TensorRT-LLM & NVIDIA GPU & AWQ GEMM 4bit & 3.42 & 5.69 & 194.86 & 6.02 \\
& & GPTQ 4bit & 3.60 & 5.88 & 202.16 & 6.08 \\
&  & INT8 & 6.53 & 8.55 & 143.57 & 5.89 \\
&   & FP16 & 12.55 & 14.61 & 83.43 & 5.85 \\ \hline
TGI & AMD GPU & AWQ GEMV 4bit & 3.62 & 36.67 & 106.84 & 6.02 \\
& NVIDIA GPU & GPTQ 4bit & 3.69 & 37.85 & 163.22 & 6.08 \\ 
 & Intel GPU & FP4 & 12.55 & 37.21 & 36.91 & 6.15 \\
& AWS Inferentia2 & NF4 & 12.55 & 37.21 & 36.32 & 6.02 \\
& & FP16 & 12.55 & 38.03 & 74.19 & 5.85 \\ \hline
MLC-LLM & NVIDIA GPU & OmniQuant 3bit & 3.2 & 5.1 & 83.4 & 6.65 \\
&  AMD GPU, & OmniQuant 4bit & 3.8 & 5.7 & 134.2 & 5.97 \\
& CPU, WebGPU, & AWQ GEMM 4bit & 3.62 & 6.50 & 23.62 & 6.02 \\
& Apple Silicon, & Q4F16 & 3.53 & 6.50 & 189.07 & - \\
& Intel GPU, & Q3F16 & 2.84 & 5.98 & 185.47 & - \\
& WASM, Adreno Mali & FP16 & 12.55 & 15.38 & 87.37 & 5.85 \\ \bottomrule
\end{tabular}}
\label{tab:engines}
\vspace{-1.5em}
\end{table*}